\documentclass[letter, 10pt, journal]{IEEEtran}      

\usepackage{graphics} 
\usepackage{epsfig} 
\usepackage{mathptmx} 
\usepackage{times} 
\usepackage{amssymb}  
\usepackage{mathtools}
\usepackage{cite}
\usepackage{url}
\usepackage{dblfloatfix}
\usepackage{color}
\usepackage{booktabs,tabularx}
\usepackage{optidef}
\usepackage[font=small,skip=3pt]{caption}
\usepackage{color,soul}

\newcommand{\robotname}{ReefFlex} 
\usepackage{siunitx}
\usepackage{subfigure}

\begin{document}

\title{\LARGE \bf{\robotname: A Generative Design Framework for Soft Robotic Grasping of Organic and Fragile objects}}

\author{\IEEEauthorblockN{Josh Pinskier\IEEEauthorrefmark{1}~\IEEEmembership{Member,~IEEE},
Sarah Baldwin\IEEEauthorrefmark{1},
Stephen Rodan\IEEEauthorrefmark{2} and
David Howard\IEEEauthorrefmark{1}~\IEEEmembership{Member,~IEEE}}\\
\IEEEauthorblockA{\IEEEauthorrefmark{1} CSIRO Robotics, Brisbane Australia}
\IEEEauthorblockA{\IEEEauthorrefmark{2} Coral Husbandry Automated Raceway Machine (CHARM) and Beyond Coral Foundation}
\thanks{Corresponding author: J. Pinskier (email: josh.pinskier{at}csiro.au).}}

\markboth{Journal of \LaTeX\ Class Files,~Vol.~, No.~, }%
{Shell \MakeLowercase{\textit{et al.}}}
%




\IEEEtitleabstractindextext{%
\begin{abstract}
Climate change, invasive species and human activities are currently damaging the world's coral reefs at unprecedented rates, threatening their vast biodiversity and fisheries, and reducing coastal protection. Solving this vast challenge requires scalable coral regeneration technologies that can breed climate-resilient species and accelerate the natural regrowth processes; actions that are impeded by the absence of safe and robust tools to handle the fragile coral. 
We investigate \textit{\robotname}, a generative soft finger design methodology that explores a diverse space of soft fingers to produce a set of candidates capable of safely grasping fragile and geometrically heterogeneous coral in a cluttered environment. Our key insight is encoding heterogeneous grasping into a reduced set of motion primitives, creating a simplified, tractable multi-objective optimisation problem. To evaluate the method, we design a soft robot for reef rehabilitation, which grows and manipulates coral in onshore aquaculture facilities for future reef out-planting. We demonstrate \textit{\robotname} increases both grasp success and grasp quality (disturbance resistance, positioning accuracy) and reduces in adverse events encountered during coral manipulation compared to reference designs. \textit{\robotname}, offers a generalisable method to design soft end-effectors for complex handling and paves a pathway towards automation in previously unachievable domains like coral handling for restoration.
\end{abstract}

}

\maketitle

\IEEEdisplaynontitleabstractindextext

%

\section{Introduction}
Coral reefs are one of the world's most important ecosystems, providing coastal protection from storms, habitat for over a quarter of the world's species and sustaining global fisheries \cite{NASA_2023}. However, the worlds coral reefs are in decline. Between 2009 and 2018, 13.5\% of the world's hard coral was lost \cite{Souter2020}, and in 2024 aerial Surveys observed bleaching in 73\% of coral reefs in Australia's Great Barrier Reef, with extreme bleaching in 39\% of surveyed reefs \cite{reef1}; trends that are only set to worsen with increasing global temperatures and human activity. Scalable robotic solutions are urgently needed to support reef rehabilitation and regeneration. These serve two purposes: to increase the volume of coral in the ocean through on-shore regeneration (coral farming) and replanting; and to selectively breed climate-resilient coral species to improve long-term survival rates. 

Coral farming facilities grow coral in onshore tanks for reef replanting. Beginning with either coral spawn or a fragment of an existing cutting, they grow the coral over months and years until it is a suitable size for replanting. In the process, they must be fed, cleaned, and moved from small to larger tanks as they grow; a task not possible with existing robotic end effectors and soft grippers, which cannot reliably grasp the geometrically diverse and easily damaged coral. A new method is needed which captures the broad range of morphologies encountered when handling natural organisms, and produces safe and reliable end-effectors.

\begin{figure}[t]
\centering
\includegraphics[height=8cm,clip]{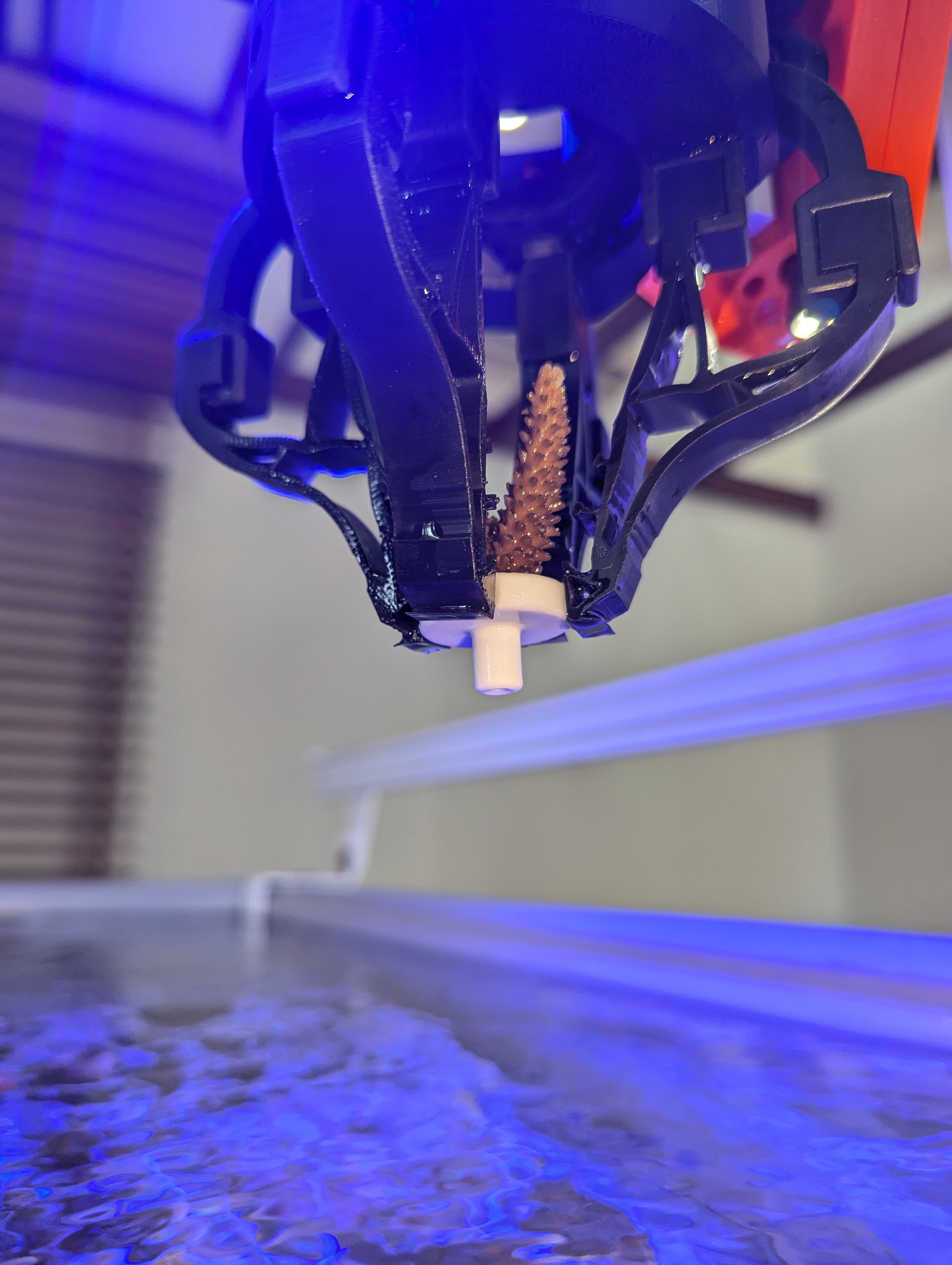}
\caption{Live Coral grasped using \robotname~at CHARM facility}
\label{fig:Heroshot}
\end{figure}


Despite substantial advances in our understanding of soft robotics, designing soft robots and soft end-effectors remains challenging. Their behaviour is dictated by a complex array of highly nonlinear physical interactions, such that even small changes in design can have substantial downstream impacts \cite{pinskier2024automated}. Further, their major advantage, embodying intelligent behaviours to control grasps through their morphology, is difficult both to quantify and to design for directly \cite{Vihmar_2023}. As a result, they are unintuitive to human designers, and there are no automated tools capable of designing them for real-world tasks \cite{Pinskier2022}. This presents a challenge in domains, such as handling living organisms, and organic and moving objects, where off-the-shelf designs are inadequate \cite{galloway2016soft}.

To create new 'intelligent' designs for these tasks, a designer (human or algorithmic) would ideally build a complete model of the grasping task, incorporating the diversity in the task's geometries (object size and shape, object substrate), orientation (relative pose of object relative to the end effector), dynamics (speed of operations, disturbances) and environmental interactions. Creating such a model is intractable in practice, so designers must rely on reduced models. For humans, this is typically an intuitive understanding of the task coupled with a baseline dataset; for algorithmic methods (computational optimisation/learning approaches) it is an idealised simulation or learned model. Hence, humans design by heuristic and computational methods create overly specialised designs for a narrow application. Whilst both are problematic on their own, the combination of intuition and explicit optimisation allows a broader search than humans can achieve whilst preventing a overfitting to non-physical simulation artifacts or hyper-specialising to a single use case. 

Several frameworks have recently emerged to design novel soft grippers and soft actuators through physics simulation. \cite{Yao2024, 10041018, doi:10.1089/soro.2023.0134, 10202189, Pinskier2024}. These produce detailed, manufacturable designs but optimise only the physical structure of the robot with little consideration for the interaction between gripper and object, a key concern for soft grasping. More general methods are required that consider a diverse set of interactions, to produce soft robots that embody intelligence and adapt to their task and environment.

To address these challenges, we develop a generative design framework for complex soft manipulation tasks, and demonstrate its capabilities in coral manipulation (Figure \ref{fig:Heroshot}).  We use a hierarchical generative design approach, which builds on our previous diversity-based topology optimisation work to find high-quality designs in a highly-nonlinear domain\cite{PinskierDiversity, 10.1145/3670693}. In this work we employ expert intuition to reduce the infinitely large space of all possible coral grasps into a tractible set of representative grasp primatives and optimise over them to generate a diverse array of designs intended to embody task performance and adaptability. The designs are generated at a high spacial resolution (small feature size) but with approximated physics, and followed by a down-selection stage where the most promising candidates are evaluated using higher accuracy simulations and experimentally. The hierarchical approach allows large sets of designs to be generated cheaply, whilst evaluating them with high-fidelity experimentation. This ensures that exploration and exploitation are balanced; and task specialisation is achieved through, rather than at the expense of embodied intelligence. 

We demonstrate \robotname's ability to generate high-performing, diverse soft grippers and translate the resulting simulated designs into real-world usage for soft manipulation. We show that we can generate high-quality, safe grasps across a set of challenging corals in a real aquaculture environment.

The key contributions of this work are:
\begin{itemize}
    \item A hybrid generative design framework for embodied intelligent grasping, which combines multi-objective soft gripping with multiple load cases
    \item A grasp benchmarking framework to assess grasp quality and grasp durability
    \item A novel compact mechanically intelligent (continuously rotatable cam-barrel) end effector
    \item A demonstration of soft grasping using the optimised designs in a real-world aquaculture environment.
\end{itemize}

Whilst we are motivated by the urgent need for automated coral farming solutions, \robotname  presents a general method for handling heterogeneous, soft and fragile goods. 

\subsection{Aquaculture Problem Definition}
To achieve reef restoration targets, coral aquaculture must be undertaken cheaply and at scale across the world, especially in developing nations with substantial coral reefs,
Hence a coral aquaculture system needs mechanically intelligent end-effectors that operate safely and reliably with minimal need for external sensing or control loop. Instead task requirements should be embodied by the design itself, with the gripper not only capable of grasping diverse corals but with inbuilt safety, such that it is physically incapable of damaging fragile coral or its own hardware. The detailed requirements are:
    
    \begin{enumerate}
        \item \textbf{Coral and substrate Grasping}: Coral are attached to cylindrical coral substrates (or plugs) for propagation (see Figure \ref{fig:Heroshot}). The gripper must be capable of grasping: (i) empty coral substrate (\SI{20}{\gram} dry mass); (ii) coral substrate with small branching coral (\num{20}-\SI{30}{\gram} dry mass, coral up to \SI{60}{\milli\meter} tall) (iii) large, diverse coral without substrate (\num{10}-\SI{30}{\gram} dry mass, up to \SI{100}{\milli\meter})
        \item \textbf{Operation in Clutter}: Coral plugs are tightly spaced on propagation trays, with grid spacing of no more than \SI{20}{\milli\meter} permitted between adjacent plugs and branching coral frequently pushing beyond the tray edges. The design must be able to operate in these confined spaces Figure~\ref{fig:Assy}(d).
        \item \textbf{Mechanically Intelligent Hardware}: Gripper must be soft enough to conform to coral shape in forward (grasping) and reverse directions (accidental collision), and prevent excessive forces being applied to the coral or gripper hardware.
        \item \textbf{Durable and Reef-Safe}: The design must be able to operate underwater, resistant to saltwater corrosion, bio-compatible, and UV stable
    \end{enumerate}
    
\section{Generative Design Framework}
Designing soft fingers to safely and reliably grasp living creatures in a cluttered environment is a challenging and unintuitive task, for which there are no mature design tools. Here we develop a generative soft design methodology for complex handling. 
The method works by decomposing soft, adaptable grasping into a set of responses to individual forces being applied along the surface of the soft gripper, such that the spectrum of coral can be tractably represented as the sum of individual load cases in a finite element simulation, and adaptive grasping becomes a low-dimensional, multi-objective optimisation problem. By explicitly considering the multiple loading conditions reflective of soft handling, our method enables the emergence of multi-functional and adaptive gripper designs through a diversity-generating multi-load topology optimisation framework.

\subsection{Diversity-Based Topology Optimisation Formulation}

Topology optimisation is an established engineering design tool, that is widely used in industry to create stiff, lightweight structures. It achieves this by finding the distribution of material that minimises the overall cost of the device given a set of physical laws and boundary conditions. Typically this is a structural mechanics problem and specified using the finite element method, but the method generalises to numerous physical (and multi-physics) systems and solver types, and has even been used to model biological growth dynamics \cite{Lowe2023, Zhang2022a}. 

We use the density-based Solid Isotropic Material with Penalisation (SIMP) topology optimisation method as the basis of this work. It defines a continuous material domain $\rho_i$ between 0 and 1 for each element in the finite element method (FEM) mesh and applies a penalty exponent $p$ to push the resulting values close to a binary value, which is necessary for manufacturing. For a detailed description of the SIMP method and its application to soft robotics see: \cite{Bendsoe2003, Pinskier2022, Chen2020}.

We define the cost function as sum of two terms, one that promotes bending of the finger via minimising $-X_{out}$ (Figure~\ref{fig:DesignDomain}) and one that minimises strain energy in the finger to strengthen the design, reduce stress concentrations, and increase the finger's lifespan:

\begin{equation}
    \phi(\rho)=\sum_{n}(wLu_n+E_n)
\end{equation}
where $n$ are the 4 or 6 load cases in the active and passive formulations described below; $u$ is the global displacement and L is a binary vector that selects relevant nodes in grasping edge; $E=u^TKu$ is the strain energy of the mechanism under loading and $K(\rho)$ is the global stiffness matrix; and $w=\num{1e5}$ is a scalar weighting. In each optimisation, a material budget is also assigned, ranging from $V_f=0.2$ to $0.5$, i.e. 20\% to 50\% of the total volume of the design domain.

As each element in the FEM mesh is an independent variable in the optimisation problem, efficient, gradient-based solvers are essential to optimise the the high-dimensional problem; whilst computationally inexpensive, gradient-based solvers are susceptible to becoming trapped in local minima, and heavily dependent on initial material distribution. We exploit this feature by using seeding diverse initial design to find numerous local minima; increasing diversity and higher quality candidates. 

\subsection{Design Domain}
\begin{figure}[t]
\centering
\includegraphics[width=1.0\linewidth]{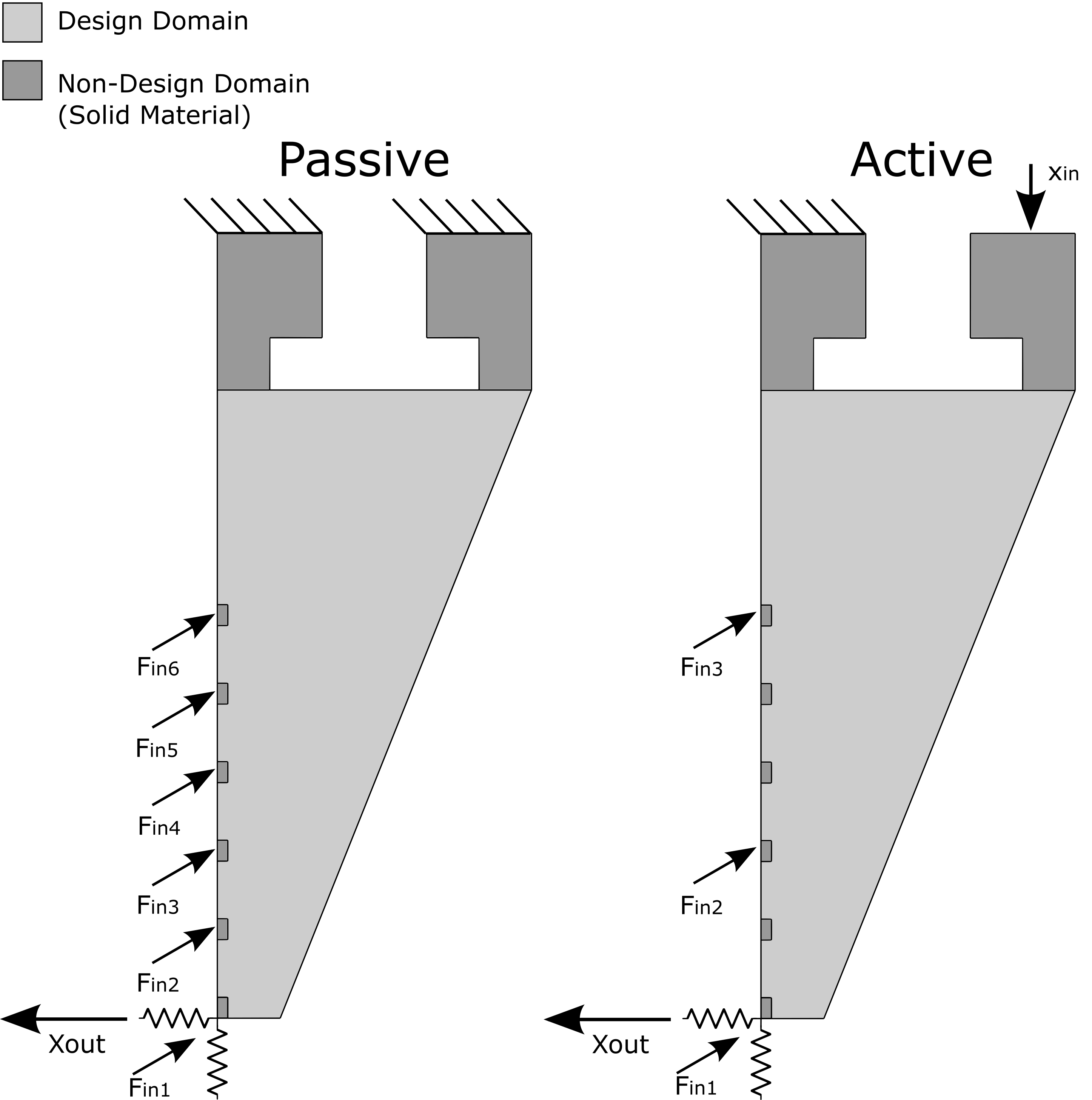}
\caption{Finger Design Domain}
\label{fig:DesignDomain}
\end{figure}

To capture soft coral grasping, we investigate two problem formulations: a passive form where the gripper is held stationary, with a set of forces applied to its contact surface to represent coral grasping; and an active form where the finger is actuated and forces are applied to the gripping surface. The two formulations are shown in Figure~\ref{fig:DesignDomain}. 

The insight for these is drawn from common Fin Ray designs, which elegantly curve around objects when enveloped, yet retain enough strength at the tip for pinch grasping. Whilst motivating, Fin Ray grippers are not well suited to handling coral and organic objects which require both pinch and power grasps. Their narrow tip and high curvature are challenging to position and can produce unstable grasps in non-smooth objects.

Both design formulations contain non-design (solid material) regions at the grippers two upper slots and six input faces where forces are applied. To increase the gripper's maneuverability amongst dense coral, the design domain tapers from a wide top to a much narrower bottom.

To represent a range of possible interactions with the coral, including grasping from the plug and picking up the coral directly, a set of 6 forces $F_{in1}$ to $F_{in6}$ are independently applied to the passive finger. The active has a reduced set of 3 forces $F_{in1}$ to $F_{in3}$ and a fixed displacement $X_{in}$. The reduced set of forces and fixed displacement biases the output towards more flexible designs with a larger rotation angle.
    
\section{Generative Design Results}
\begin{figure*}[t]
\centering
\includegraphics[width=1.0\linewidth]{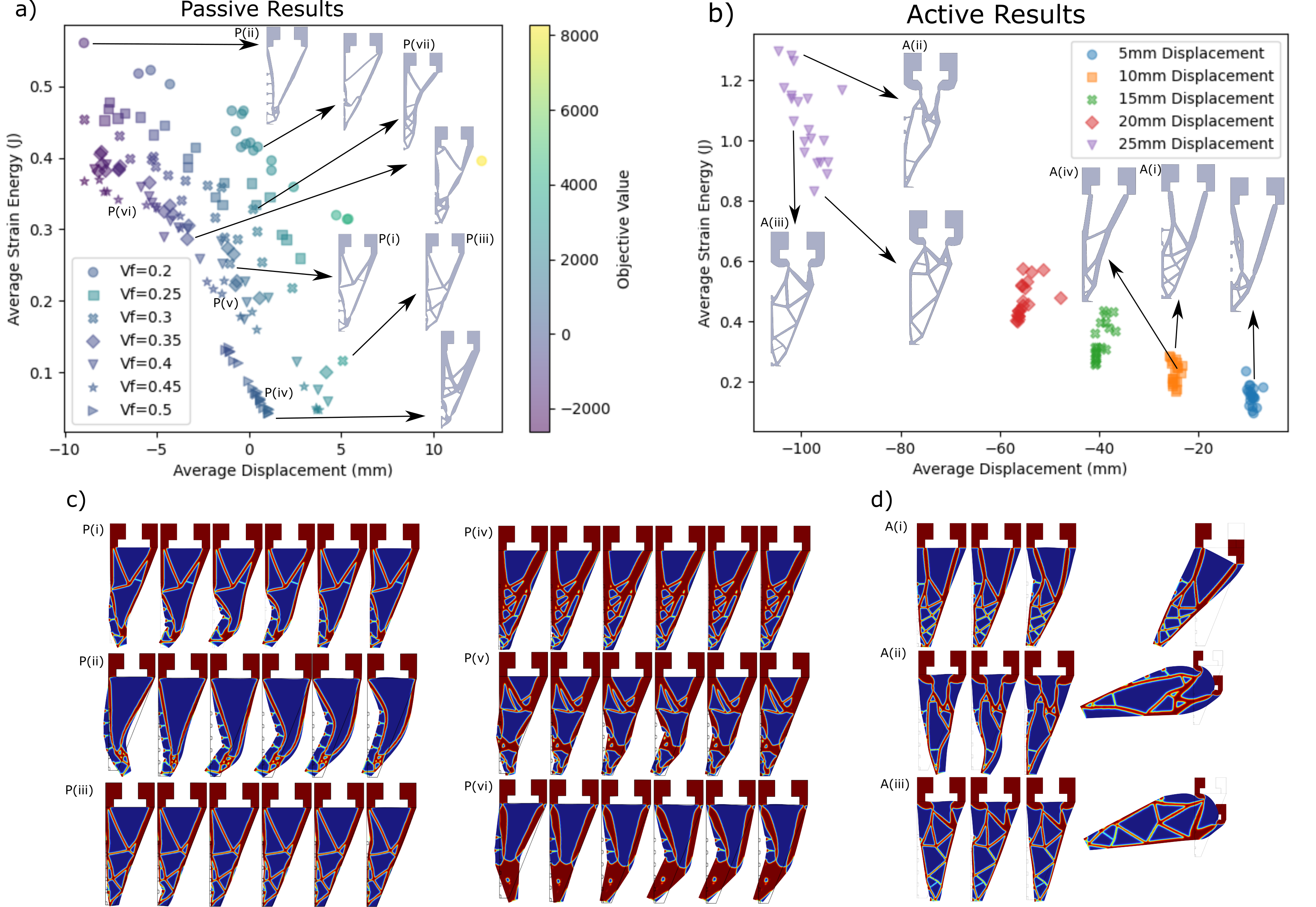}
\caption{a) Complete numerical results of Passive Finger Optimisations illustrating Pareto front of the designs and trade-off between increasing displacement and reducing strain. b) Complete numerical results of active Finger Optimisations. c) Visualisation of representative selection of fingers under all six load cases (Left to Right: $F_{in1}$ to $F_{in6}$), where red is solid material and blue is void. d) Visualisation of selected fingers under four load cases (Left to Right: $F_{in1}$, $F_{in2}$, $F_{in3}$, $X_{in}$).}
\label{fig:Combined_results}
\end{figure*}

\begin{figure}[t]
\centering
\includegraphics[width=1.0\linewidth]{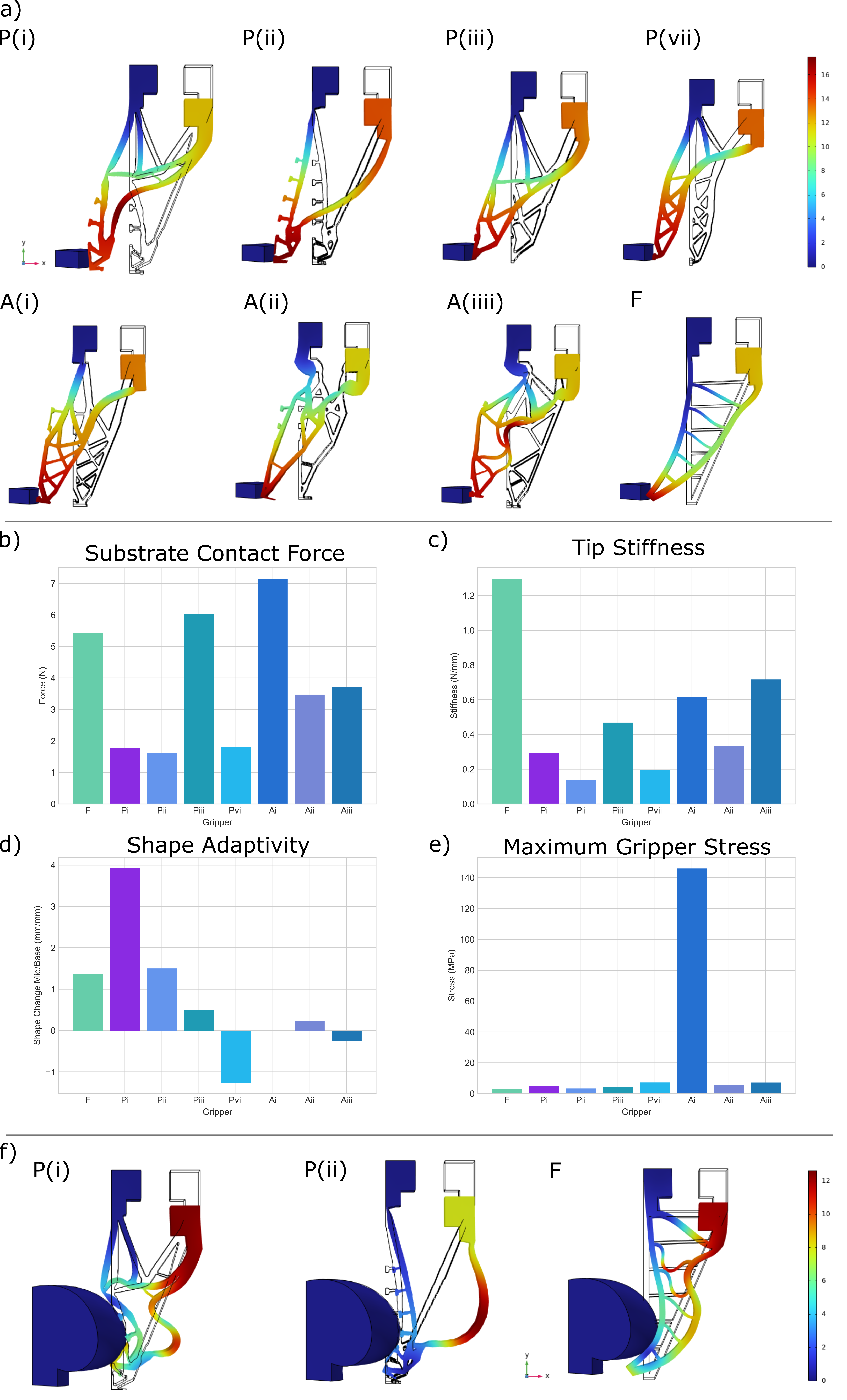}
\caption{a) Simulated deformation profiles for selected optimised grippers when grasping fixed coral substrate. b-e) Grasp Performance Validation using Non-linear contact model with substrate. b) Contact force between gripper and substrate at full extension, c) Gripper Tip Stiffness, d) Shape adaptivity, the relative change in displacement between the gripper middle and tip when a force is applied at the middle of its contact face ($F_{in3}$), e) Maximum stress experienced by the gripper during substrate grasping. f) Illustration of shape adaptability of different grippers when grasping a simplified brain coral. Gripper $P(i)$ is able to adapt and conform to the coral geometry far better than $P(ii)$, the benchmark Fin Ray, or other optimised designs }
\label{fig:ComsolVerification}
\end{figure}

\subsection{Passive Optimisation}
A total of 140 optimisations runs were evaluated for the passive problem, comprising 20 runs of different initial material distributions at each of $V_f=\{0.2,~0.25,~0.3,~0.35,~0.4,~0.45,~0.5\}$. 
The combined results of these optimisations are presented in Figure~\ref{fig:Combined_results}(a), alongside a representative set of soft grippers under the six loading conditions (Figure~\ref{fig:Combined_results}(b)). It is expected that increasing the volume of material in the design will bias the results towards stiffer designs with a lower strain energy and smaller absolute displacement (i.e. the bottom right corner of the plot), whilst lower material will result in more flexible designs (top left of the plot). 
The ideal situation is a design in the bottom left, which exhibits high flexibility and adaptability, whilst retaining enough stiffness to apply sufficient force and without placing large strains on the material.

Higher volume fractions generally result in lower strain energies, however smaller volume fractions do not necessarily result in larger magnitude of displacement (i.e more negative); instead the Pareto front is formed exclusively by the designs with $V_f=\{0.4,~0.45,~0.5\}$. Numerically, this makes sense, as the volume fraction gives an upper bound on material, lower volume fractions are only able to search a subset of the design space of higher volume fractions. However, it is frequently the case that valid designs cannot be found that span the Pareto front.
This effect is evident in the $V_f=0.5$ results, which are tightly clustered together in a line close to zero average displacement. These results are heavily biased towards low strain energy and converged on relatively stiff designs which restricts both beneficial and parasitic displacements. In contrast, every other run gave a spread of results across displacement and strain energies.

The higher volume runs were able to produce designs which, at least measured by the objective value, are unambiguously better than the lower ones. Whilst this numerical result is an imperfect abstraction of the true problem, it is nevertheless instructive. Higher permitted volumes create soft fingers that resemble traditional compliant mechanisms. That is, their deformation is lumped in a small number of compliant hinges, with relatively thick sections between them. Reducing the material budget, prevents the formation of these thick members, forcing compliance to be distributed throughout the finger and giving a more flexible and adaptable design, akin to benchmark soft robotic designs. 
This effect is evident when comparing designs P(ii) and P(iv) in Figure~\ref{fig:Combined_results}(a). The two produce comparable average displacements, but
P(ii) does so through a smooth curvature well suited to irregular shape adaptation, whilst P(iv) rotates only about two compliant hinges. These large displacement, low stiffness designs can be strengthened through the inclusion of additional linkages. For example P(i) is similar to P(ii), with additional truss-like support at the base, which localises the bending closer to the tip. Designs P(v) is similar to P(vi) and exhibits the same effect. At the extreme end of the design space, fingers converge towards rigid trusses, which minimise cost by reducing displacement (and hence strain) as much as possible. Designs P(iii) and P(iv) illustrate this at different volumes, with the exception of small hinges at the very tip, they are configured to resist forces, rather than conform to them.

\subsection{Active Optimisation}
In the active problem, the volume fraction was fixed at $0.3$, with the input displacement instead used as the swept variable. Here, 100 optimisation runs were evaluated, comprising 20 runs with different initial material distributions at each of $X_{in}=\{5,~10,~15,~20,~25\}\si{\milli\meter}$. The combined results of these optimisations are presented in Figure~\ref{fig:Combined_results}(c), alongside a representative set of soft grippers under the four loading conditions (Figure~\ref{fig:Combined_results}(d)). Whilst the 5 sets of designs are each tightly clustered, the trade-off between objectives remains evident as the higher displacement designs must absorb more strain energy. 
The major benefit of this problem configuration is that the finger's bending is modelled rather than implied, giving a more realistic representation of the scenario. However, the relatively large displacement created by the bending tends to overwhelm the bending created by contact with objects, reducing the incentive to optimise for adaptability. This is evident in Fingers A(i) and A(iii), which have compliant hinges and a lever arm near the top supports, but otherwise resemble structural trusses..

A relatively large spread of results is seen within the \SI{25}{\milli\meter} optimisation run, giving some designs which prioritise stiffness, and others which are more compliant. This is especially evident in Finger $A(ii)$, where a large opening in its centre that permits tip bending. This localised bending increases overall displacement only minimally, but it is expected to improve grasp quality in the uncertain coral environment.

\section{End effector Hardware Design} \label{method}
\begin{figure*}[t]
\centering
\includegraphics[width=1.0\linewidth]{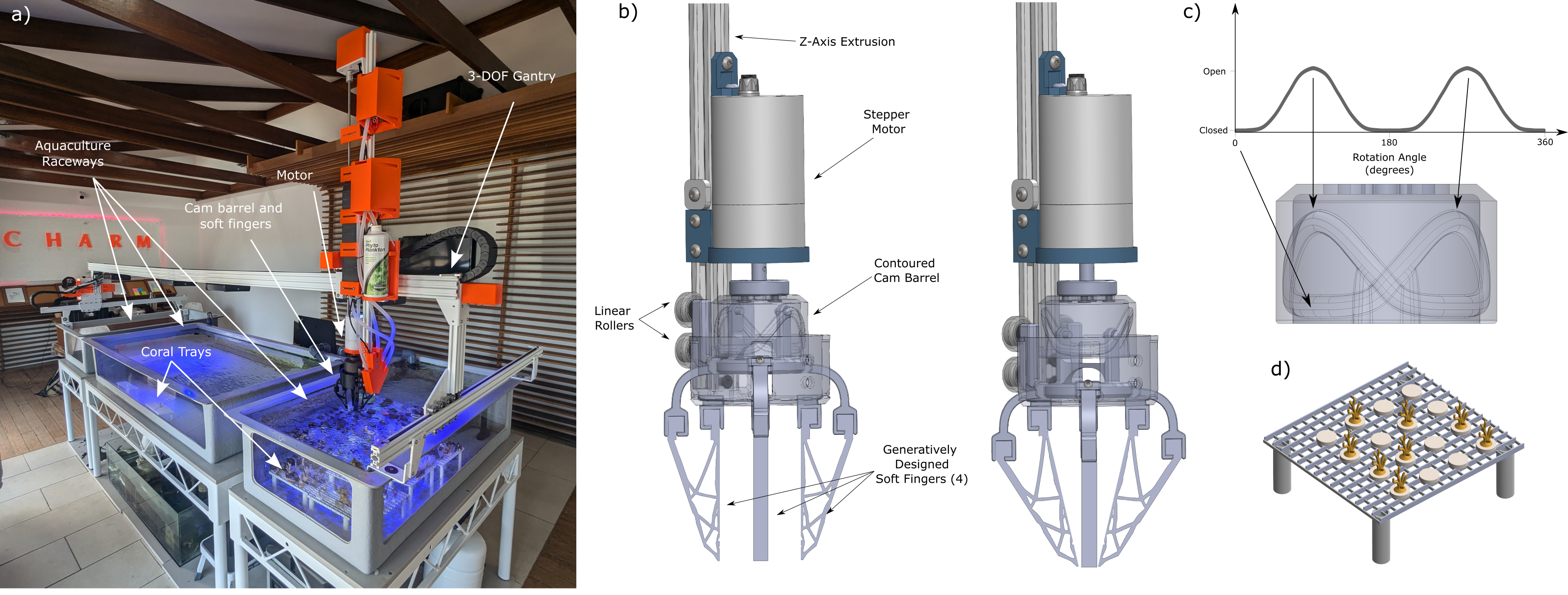}
\caption{(a) CHARM Robot and Raceway system, (b) \robotname Gripper assembly, showing major design components. Left: In open position, Right: closed position. c) Cam profile and corresponding barrel positions. d) Coral propagation tray with empty and coral containing plugs}
\label{fig:Assy}
\end{figure*}
To experimentally evaluate the generated soft fingers and integrate them into commercial operations, we designed a mechanically intelligent hand, that comprises four main components:  (1) A stepper motor, (2) contoured cam barrel, and (3) linear sliders which together create the grasping motion profile; and (4) the generatively designed soft fingers that interact with coral. The hand forms part of the CHARM robotic coral farming system, which additionally contains a 3-DOF gantry, aquaculture raceways (tanks), and water monitoring and filtration systems. Both the CHARM system and \robotname~ are shown in Figure~\ref{fig:Assy}. CHARM aims to provide a low-cost aquaculture system that can be distributed around the world to local reef-restoration organisations. 

The contoured cam barrel produces a smooth cyclical motion trajectory for the fingers.
As the cam rotates, two internal pins are driven vertically by the cam's internal contours, moving the outer arm of each soft finger in turn and causing them to close inwards (Figures \ref{fig:Assy}(b) and \ref{fig:Assy}(c)). The output stage is supported by four rollers, such that it is restricted to moving only in the vertical axis.
The circular cam profile prevents excessive forces from being applied to the coral and does so without the need for mechanical stops. 

The vast majority of the mechanism is 3D printed, with the fingers being printed in Figure 4 Rubber-65A; and the cam barrel, rollers, housing and linkages in Figure 4 Pro-Blk 10.
To simplify connection and interchange, the soft fingers are simply press-fit into their fixtures. 

Four fingers are used in the design, with fingers oriented at $\pm\SI{45}{\degree}$ and $\pm\SI{135}{\degree}$ from the coral tray's vertical axis. As the coral's grid arrangement is aligned with the trays, this maximises the area that fingers can sweep through without contacting adjacent corals whilst grasping.

\section{Experimental Results \& Discussion} \label{experiments}
\begin{figure*}[t]
\centering
\includegraphics[width=1.0\linewidth]{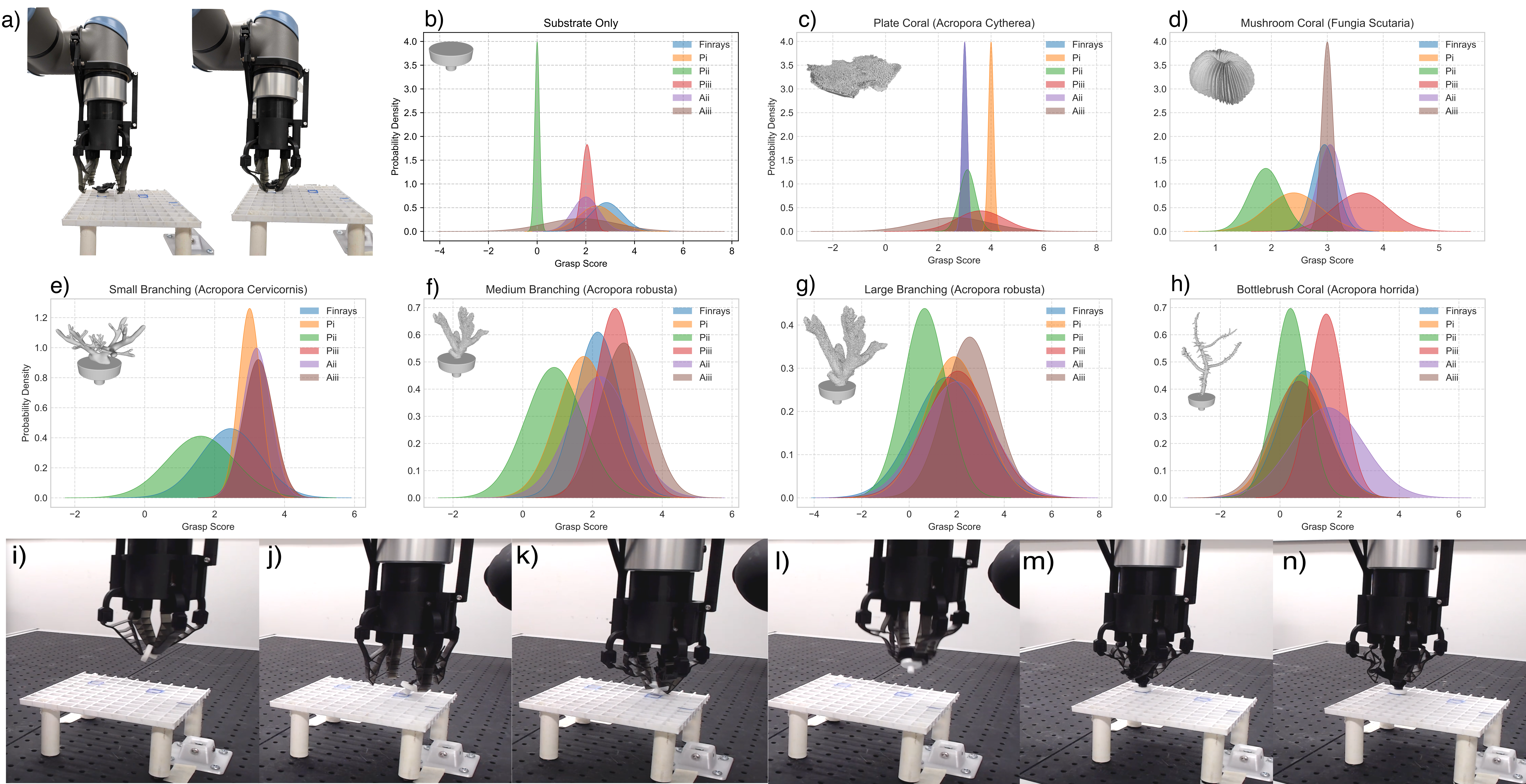}
\caption{a) Experimental grasp reliability and durability testing using UR10 arm. Each object is grasped and released before being regrasped and shaken at 3 different speeds. b-h) Grasp quality results for 6 optimised finger designs (5 optimised and Fin Ray benchmark) across 7 objects (empty substrate and 6 corals). Grasp scores correspond to: 
i-n) Grasp failure modes and challenges identified during experimental testing: i) object dropped due to in-hand slippage caused by initial object misalignment, j) failed object release due to object-gripper entanglement, k) failed release due to in-hand object rotation, l) dynamic slippage due to finger movement induced by rapid acceleration, m) torsional finger reconfiguration created by protruding coral branches, n) large deformation asymmetric grasping induced by protruding coral branches. Note (m) and (n) successfully grasped and released}
\label{fig:UR10Results}
\end{figure*}

\subsection{Finger Selection}
To validate these results experimentally, a set of 7 fingers from the active and passive optimisations were selected for experimental testing. The fingers were selected based on their perceived capacity to: (i) produce sufficient bending to reach coral plugs, (ii) generate a sufficiently strong pinch grasp to hold coral plugs, (iii) conform to coral during grasping, (iv) represent a cross-section of the optimised designs.

Although seemingly innocuous, the first criterion ruled out the very stiff (low strain energy) finger designs produced by the passive optimisation. The large forces required to drive these fingers exceeds the limits of the cam barrel and would cause it to stall (e.g P(iv) and P(v)). The second criterion removes very flexible designs (e.g P(ii)).

The final designs selected were P(i), P(iii), P(vii) and A(i), A(ii) and A(iii). A Fin Ray soft finger and P(ii) were also tested for reference.

\subsection{Grasp Validation in Simulation}
The selected designs were first assessed in a high-fidelity finite element simulation to evaluate their ability to deform sufficiently to grasp and hold an empty substrate plug, and to conform to different coral configurations.
In contrast to the linearised topology optimisation model, here the designs are evaluated using a geometrically nonlinear simulation in COMSOL multiphysics. The nonlinear model captures the large displacements experienced during grasping.
Each gripper is evaluated in two tests: a free displacement test, and a substrate contact test. Both tests are conducted on a single soft finger with the results mirrored in two planes to reflect the 4 fingered grasping of the final design.
In the first test, the soft gripper is driven to its grasp position, before a \SI{1}{\newton} force is applied in the x-direction at its tip. The force is then removed and a second \SI{1}{\newton} force is applied midway up its contact face, at $F_{in3}$.
The second test simulates grasping an empty coral substrate, in it the gripper is driven from rest into its grasp position with a substrate plug placed at the center of the four fingers. Both assume a linear isotropic material with $E=\SI{23}{\mega\pascal}$, reflective of the Rubber 65A material.

The results of the two tests are summarised in Figure \ref{fig:ComsolVerification}, with the results of the second test visualised in Figure \ref{fig:ComsolVerification}(a), and summary statistics in Figure \ref{fig:ComsolVerification}(b-e).
As expected, high strain energy translates into a relatively stiff fingertip (Figure \ref{fig:ComsolVerification}(c) and hence the gripper can exert a large force on the substrate (Figure \ref{fig:ComsolVerification}(b)). In order of SE, gripper P(iii) was lowest followed by P(i), P(vii) and P(ii). Although the magnitudes are disproportionate, their ordering is reflected in their ranking in tip stiffness and contact force. The active designs exhibit a similar pattern, with lower SE translating to higher contact forces (although not necessarily higher tip stiffness). High stiffness and contact force is a desirable feature as it enables large corals to be grasped from the substrate, but it can come at the expense of softness, which is a critical element in adaptable grasping, and induces high stresses in the gripper.
We quantify shape adaptivity as the difference in deformation between $F_{in3}$ and the tip when a $\SI{1}{\newton}$ force is applied at $F_{in3}$ (passive formulation) or $F_{in2}$ (active).
A larger relative difference indicates greater ability to conform to objects. Here, softer designs excel, notable $P(i)$ whose tip moves upwards when its centre moves inwards, allowing it to wrap around small objects (illustrated in Figure \ref{fig:ComsolVerification}(f)). 
All grippers tested have relatively low internal stresses, with the notable exception of $A(i)$, which has a large stress concentration caused by a thin notch near its actuated side, making it likely to fail during use. Hence, it is removed from further testing.

\subsection{Experimental Grasp Quality Testing}
The remaining 6 designs were then tested experimentally in a dry-testing environment with 7 different objects, six 3D printed corals (printed in PLA) \cite{siDigitization,thingiverseCryohabitatCoral} glued to a substrate, and an empty substrate. Whilst not fully representative of the final use case, this environment allows a greater degree of control and repeatability over the testing than the true aquaculture environment.
An industrial robot arm (UR10e) is used to perform pick and place testing, and impulse response (shake testing), assessing both the grasp reliability and its resistance to impulses.
Whilst grasp success rate is typically used to measure grasp quality in robotics, grasp robustness (sensitivity of the grasp to the object pose) and impulse response are also critical features for ongoing use. Impulse response dictates the ability to securely and stably hold an object when moving over uneven terrain, as well as the maximum grasp rate in industrial applications, where it is necessary for the end effector to quickly (i.e with high acceleration) move from one point to another and back.
Each test is conducted as follows: for each gripper/object pairing the gripper first grasps the object from a randomised pose within a starting cell on the coral tray, raises it, and moves it to a secondary cell, before releasing (start and end cells are shown in blue in Figure \ref{fig:UR10Results} (a)). It is then regrasped and raised before being moved through a square trajectory (in the horizontal plane) with $\SI{10}{\milli\meter}$ side lengths, at three different acceleration rates ($\SI{128}{\degree\per\second}$, $\SI{256}{\degree\per\second}$, and $\SI{1024}{\degree\per\second}$). The object moves through the square 5 times at each speed before advancing to the higher speed. At the end of process the object, if still held, is returned to the second point. To reset the test, the object is manually placed back in the initial grid cell with a randomised pose.
The test is repeated 20 times for each gripper/object pair allowing the grasp robustness to be evaluated over a set of poses.
The tests are then scored as: 0 = failed grasp, 1 = successful pick and place, 2 = object held during low speed shaking, 3 = held during medium shaking, 4 = held during high speed shaking and placed back into tray.

\subsubsection*{Grasp Results}
The resulting grasp scores are fitted to a normal distribution and presented in Figure \ref{fig:UR10Results}(b-h).
Looking across the seven plots, it is evident that gripper $P(ii)$ performs worst of the 6 tested. It was unable to grasp the empty substrate in any of the 20 tests conducted (fingers slipped off the substrate without lifting it), and was the worst performing gripper in all but one of the remaining objects. The shape adaptability of $P(i)$ generated extremely durable grasps for small, light objects, notably it survived rapid shaking in all 20 of the tests of the plate coral. However it was less reliable with heavier objects, as it was unable to withstand the larger impulse forces generated. In contrast, the larger displacement grippers $A(ii)$ and $A(iii)$ were very successful with the larger corals (small, medium and large branching and bottlebrush), as they could leverage this large movement to successfully reconfigure to multiple stable grasp modes including torsional grasping and asymmmetric grasping in the presence of protruding features (Figure \ref{fig:UR10Results}(m-n). In other designs these would cause a grasp failure as the fingers would bump the object or it would slip in-hand (Figure \ref{fig:UR10Results}(k). However, they were relatively poor at grasping smaller objects, as the elastic energy stored in the grippers created unstable grasps. Whilst the benchmark Fin Ray design performed reliably across the spectrum of objects, $P(iii)$ was the best performing overall, it gave a stable grasp which ensured a secure hold across the objects.

\subsection{In-Situ Demonstration}
\begin{figure}[t]
\centering
\includegraphics[width=1.0\linewidth]{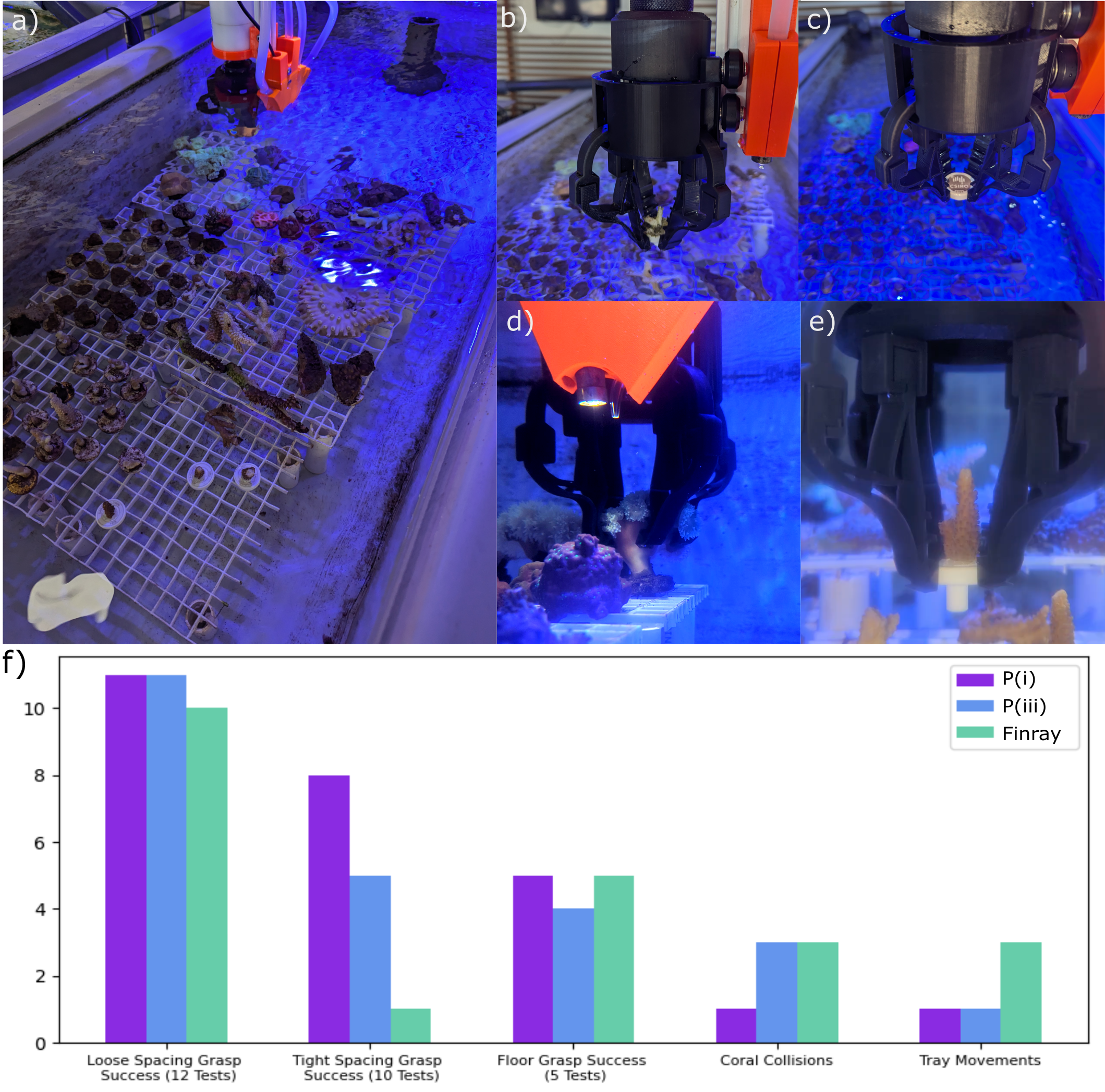}
\caption{Selection of successful grasped items during in-situ testing: a) Example raceway configuration,  b) small branching coral partially grasped by coral, c) empty 'CSIRO' coral substrate, d) Large soft coral, e) Small branching coral held by substrate}
\label{fig:InSitu-Results}
\end{figure}
From the previous test, three designs were selected for in-situ evaluation at the CHARM aquaculture facility, $P(i)$ (the most adaptable), $P(iii)$ (the best experimentally) and the benchmark Fin Ray.
The performance of the three designs was experimentally investigated in a series of 3 tests, which are representative of the real-world aquaculture usage. Firstly, 12 coral plugs (a mix of empty and with coral) were setup on a tray with adjacent coral with 2 grid cells between each plug, reflective of standard usage, the 12 were to be collected one at a time and moved to an empty tray in the same tank. Secondly, 10 coral were tightly spaced, one grid cell apart and the test repeated, reflective of a cluttered aquaculture environment. Finally, 5 coral were placed on the bottom of the raceway to be grasped and returned to a tray, to record recovery from a failed grasp. Adverse events were recorded during testing: the gripper colliding with a coral outside of the grasping process was considered minor adverse event, and the gripper displacing a coral tray through a coral collision a major event. The CHARM robot was teleoperated by an experienced operator during these tests.

The results of these tests are presented in Figure~\ref{fig:InSitu-Results}. Whilst all three designs performed well in loose spacing and floor grasping, finger P(i) excelled in tight spacing, successfully grasping 8/10 coral plugs. Its small deformation and large contact surface produced a durable grasp and greatly outperformed the Fin Ray (1/10 success), which was unable to produce reliable contact because of its small contact surface. Finger P(i), similarly gave the fewest adverse events.

\section{Conclusion} 
\label{conclusion} 

ReefFlex presents a novel generative design framework for soft robotic grasping, enabling safe and adaptive manipulation of fragile and geometrically diverse objects. By integrating expert intuition with multi-objective topology optimisation, ReefFlex produces mechanically intelligent grippers that outperform traditional designs in both simulation and real-world aquaculture environments. Experimental validation across robotic and in-situ testing demonstrates improved grasp quality and robustness, highlighting its value in complex handling and utility for scalable coral farming. Beyond coral restoration, ReefFlex offers a generalisable approach for designing soft end-effectors in complex, cluttered, and delicate handling tasks, paving the way for automation in domains previously inaccessible to robotics.


\section*{Acknowledgement}
This work is supported by the Science and Industry Endowment Fund, CHARM and CSIRO Kickstarter. Australian Design \#202510295 has been registered by CHARM IP Pty Ltd.

\bibliographystyle{ieeetr}
\bibliography{root}

\begin{IEEEbiography}[{\includegraphics[width=1in,height=1.25in,clip,keepaspectratio]{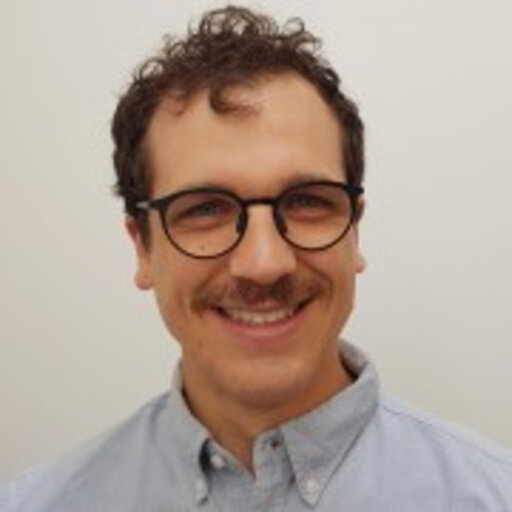}}]{Josh Pinskier}
is a senior research scientist with CSIRO Robotics. His research investigates the interaction between soft robots and the environment and explores of closed-loop design processes to artificially synthesise unconventional, lifelike and high-performing soft robots for applications including environmental resoration and agriculture. He has particular expertise in topology optimisation, evolutionary optimisation, soft robotic manipulation and multi-material fabrication. He received a Bachelor of Mechatronics Engineering with Honours and Bachelor of Commerce from Monash University in 2015 and a PhD in Compliant Mechanism Synthesis in 2019. 
\end{IEEEbiography}

\begin{IEEEbiography}[{\includegraphics[width=1in,height=1.25in,clip,keepaspectratio]{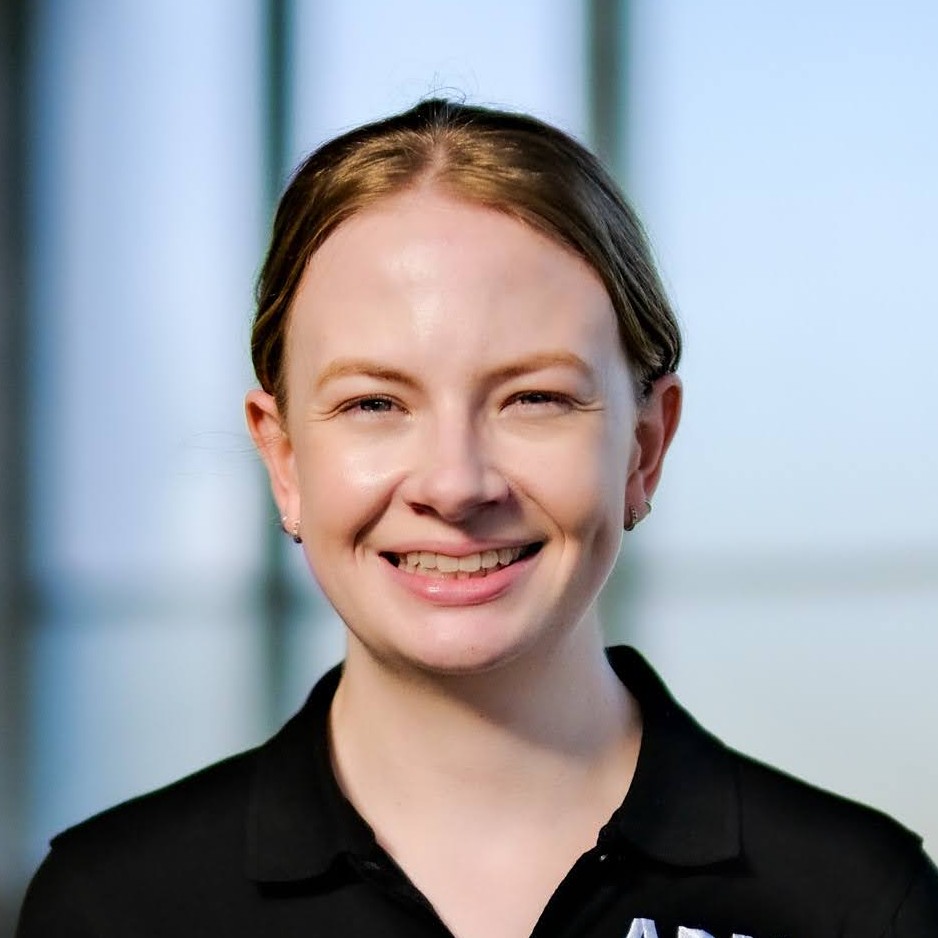}}]{Sarah Baldwin}
is a mechatronics engineer specialising in advanced manufacturing and soft robotics. Her work spans cross-disciplinary research in areas such as coral restoration technologies, space hardware, and novel soft gripper design. She has co-authored several publications and presented at international robotics conferences, including RoboSoft. Sarah completed her honours thesis in computer vision for 3D printing failure detection and now focuses on translating research principles into practical, industry-focused automation and manufacturing solutions
\end{IEEEbiography}

\begin{IEEEbiography}[{\includegraphics[width=1in,height=1.25in,clip,keepaspectratio]{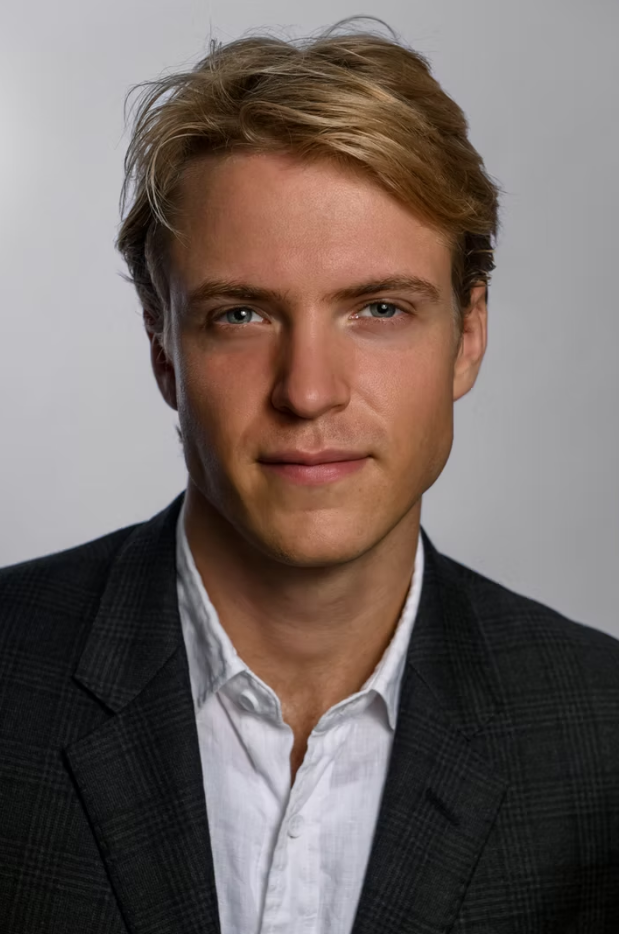}}]{Stephen Rodan}
Mechanical engineer and nuclear physicist with an aerospace engineering background. Professional expertise in aquaculture, robotics, and renewable energy. Inventor of the Coral Husbandry Automated Raceway Machine (CHARM) and founder of not-for-profit Beyond Coral Foundation. Entrepreneur, private contractor, and philanthropist advancing sustainable island communities, biomechatronics, and sustainable marine restoration.
\end{IEEEbiography}

\begin{IEEEbiography}[{\includegraphics[width=1in,height=1.25in,clip,keepaspectratio]{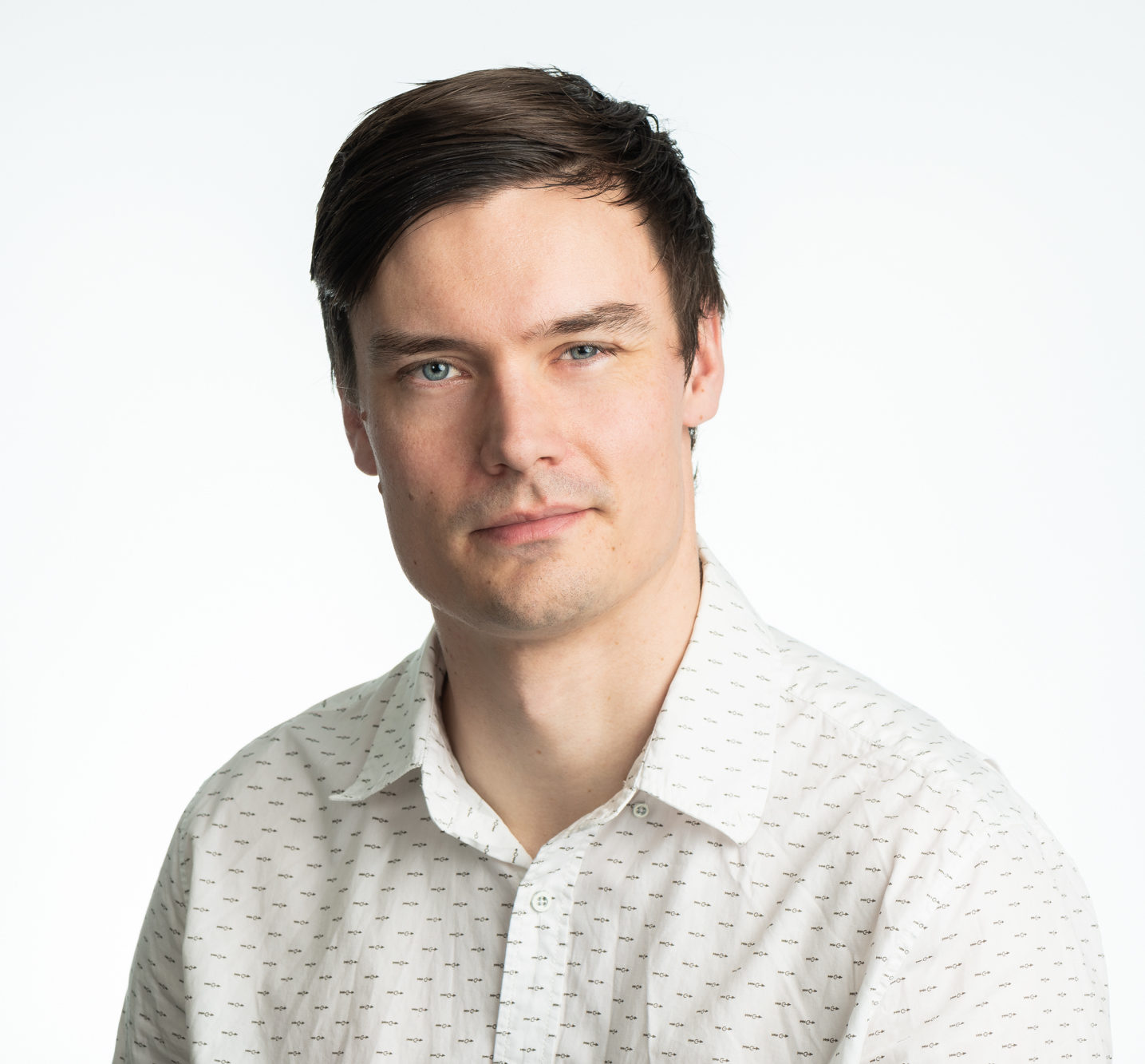}}]{David Howard} leads the Robotic Design and Interaction Group and is a Principal Research Scientist in the Cyber Physical Systems program at CSIRO, Australia's national science body. He leads multiple projects at the intersection of soft robotics, evolutionary machine learning, and the computational design of novel physical objects. He currently leads the AI4Design portfolio. His interests include nature-inspired algorithms, learning, autonomy, soft robotics, the reality gap, and evolution of form. His work has been featured in local and national media.

He received his BSc in Computing from the University of Leeds in 2005, and the MSc in Cognitive Systems at the same institution in 2006. In 2011 he received his PhD from the University of the West of England. He is a member of the IEEE and ACM, and an avid proponent of education, STEM, and outreach activities. His work has been published in IEEE and Nature journals.

\end{IEEEbiography}

\end{document}